\documentclass[runningheads]{llncs}
\usepackage{graphicx}
\newif\ifsubmit

\usepackage[T1]{fontenc}
\usepackage{amsmath}
\usepackage{amssymb}
\usepackage{amsfonts}
\usepackage[english]{babel}
\usepackage{graphicx}
\usepackage{subcaption} 
\usepackage[usenames,dvipsnames]{xcolor}
\usepackage{xspace}
\usepackage[linesnumbered]{algorithm2e}
\usepackage{paralist}
\usepackage{booktabs}
\usepackage{longtable}

\usepackage{diagbox}

\usepackage[pdfpagelabels=true]{hyperref}
\usepackage{color}

\usepackage{cleveref}

\crefname{equation}{Eq.}{Eq.}
\crefname{pluralequation}{Eqs.}{Eqs.}

\crefname{figure}{Fig.}{Fig.}
\crefname{pluralfigure}{Figs.}{Figs.}

\crefname{section}{Sect.}{Sect.}
\crefname{pluralsection}{Sects.}{Sects.}

\crefname{appendix}{App.}{App.}
\crefname{pluralappendix}{Apps.}{Apps.}

\crefname{table}{Tab.}{Tab.}
\crefname{pluraltable}{Tabs.}{Tabs.}

\crefname{definition}{Def.}{Def.}
\crefname{pluraldefinition}{Defs.}{Defs.}

\crefname{theorem}{Theorem}{Theorems}
\crefname{pluraltheorem}{Theorems}{Theorems}

\crefname{lemma}{Lemma}{Lemma}
\crefname{plurallemma}{Lemmas}{Lemmas}

\crefname{example}{Example}{Example}
\crefname{pluralexample}{Examples}{Examples}

\crefname{assumption}{Assumption}{Assumption}
\crefname{pluralassumption}{Assumptions}{Assumptions}

\crefname{remark}{Remark}{Remark}
\crefname{pluralremark}{Remarks}{Remarks}

\usepackage{tikz}
\usepackage{pgfplots}
\usepackage{pgfplotstable}
\pgfplotsset{compat=1.17}
\usetikzlibrary{shapes.gates.logic.US,shapes.geometric,positioning,fit,calc}
\usepgfplotslibrary{colorbrewer} 

\makeatletter
\def\orcidID#1{\smash{\href{https://orcid.org/#1}{\protect\raisebox{-1.25pt}{\protect\includegraphics{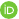}}}}}
\makeatother

\newcommand{\colorpar}[3]{\colorbox{#1}{\parbox{#2}{#3}}}
\newcommand{\marginremark}[3]{\marginpar{\colorpar{#2}{3.5em}{\color{#1}#3}}}

\ifsubmit
  \newcommand{\mv}[1]{}
  \newcommand{\ms}[1]{}
  \newcommand{\ml}[1]{}
  \newcommand{\mh}[1]{}
\else
  \newcommand{\mv}[1]{\marginremark{black}{yellow}{\tiny{[MV]~ #1}}}
  \newcommand{\ms}[1]{\marginremark{purple}{white}{\tiny{[MS]~ #1}}}
  \newcommand{\ml}[1]{\marginremark{black}{red}{\tiny{[MLZ]~ #1}}}
  \newcommand{\mh}[1]{\marginremark{white}{blue}{\tiny{[MH]~#1}}}
\fi

\begin{document}
\title{Fuzzy Fault Trees Formalized\thanks{This research has been partially funded   by ERC Consolidator grant 864075 CAESAR and the European Union’s Horizon 2020 research and innovation programme under the Marie Skłodowska-Curie grant agreement No. 101008233.}}
%
%
\author{Thi Kim Nhung Dang\inst{1}\orcidID{0000-0002-3235-5952} \and
Milan Lopuhaä-Zwakenberg\inst{1}\orcidID{0000-0001-5687-854X} \and
Mariëlle Stoelinga\inst{1,2}\orcidID{0000-0001-6793-8165}}

\authorrunning{Dang et al.}
%
\institute{University of Twente, Enschede, the Netherlands  \\
\email{\{t.k.n.dang, m.a.lopuhaa, m.i.a.stoelinga\}@utwente.nl}\\
\and Radboud University, Nijmegen, the Netherlands \\
\email{m.stoelinga@cs.ru.nl}\\
}

\allowdisplaybreaks

\maketitle              
\begin{abstract}
Fault tree analysis is a vital method of assessing safety risks. It helps to identify potential causes of accidents, assess their likelihood and severity, and suggest preventive measures. Quantitative analysis of fault trees is often done via the \emph{dependability} metrics that compute the system's failure behaviour over time. 

However, the lack of precise data is a major obstacle to quantitative analysis, and so to reliability analysis. Fuzzy logic is a popular framework for dealing with ambiguous values and has applications in many domains. A number of fuzzy approaches have been proposed to fault tree analysis, but 
---to the best of our knowledge---
none of them provide rigorous definitions or algorithms for computing fuzzy unreliability values.

In this paper, we define a rigorous framework for fuzzy unreliability values. In addition, we provide a bottom-up algorithm to efficiently calculate fuzzy reliability for a system. The algorithm incorporates the concept of $\alpha$-cuts method. That is, performing binary algebraic operations on intervals on horizontally discretised $\alpha$-cut representations of fuzzy numbers. The method preserves the nonlinearity of fuzzy unreliability. Finally, we illustrate the results obtained from two case studies.

\keywords{Fault trees \and reliability analysis \and fuzzy numbers}
\end{abstract}
\section{Introduction}
\label{sec:intro}
\noindent\emph{Fault trees.} Fault trees (FTs) are a systematic, graphical tool used to identify potential sources of failure in a system. By analyzing an FT, a designer can identify potential failure paths, understand the impact of failures, and determine the best measures to prevent them. A FT is a directed acyclic graph (DAG) that consists of basic events (BEs) representing atomic failure events, intermediate AND/OR gates whose activation depends on that of their children, and the root representing system failure. The system fails when the root is activated. We provide a brief overview of FTs in Sec.\ref{sec:FTs}

\noindent\emph{Quantitative analysis.} Whereas qualitative analysis identifies critical factors and root causes in a system design, quantitative analysis generates relevant numerical values. Typical metrics include the system reliability, availability, mean time to failure, etc. We are focused on \emph{(un)reliability}, a method that calculates system failure probability, where BEs are equipped with probabilistic information. Unreliability is a critical safety metric to ensure system safety and availability.    

\noindent\emph{Uncertain parameters.} Conventional fault tree analysis (FTA) assumes failure probability is exact or a single estimated value. However, it is infeasible in practice to estimate a failure rate or probability as we must handle two types of uncertainty: aleatoric uncertainty that stems from natural fluctuations, modeled as probability distributions; and epistemic uncertainty that stems from a lack of knowledge, in our case not knowing exact values for failure rates. Thus, methods that allow us to deal with uncertain parameter values are urgently needed.

\noindent\emph{Fuzzy theory.} Fuzzy theory is a common framework for handling imprecise data/uncertainty. Fuzzy logic has been applied in various domains such as control systems \cite{deBarros2017afirst}, medical imaging, economic risk assessment, decision trees \cite{Basiura2015advances} and machine learning \cite{couso2019fuzzy}. Fuzzy logic models uncertainty/similarity by a range of values, and to each of these a possibility value in $[0,1]$ is assigned by means of a \emph{membership function}. Fundamentals of fuzzy theory are summarised in Sec.\ref{sec:fundamentals_fuzzy} and based on our earlier paper \cite{dang2024fuzzy}.

While fuzzy theory has been proposed in many studies to overcome the limitation of conventional FTA \cite{mahmood2013FFTA,tanaka1983fault,lin1997hybridFTA,singer1990fuzzysetFT}, it has been lacking in mathematical rigor in much of the earlier work. In addition, there are no efficient algorithms for calculating unreliability with fuzzy parameters. Performing exact calculations for nonlinear operations is computationally expensive and difficult to implement in practice. As a consequence, there are approximations that result in fuzzy numbers of the same type as the operands \cite{tanaka1983fault}. Thus, they do not provide the right-shaped resulting number. Moreover, if the operands are of two different types, the approximation is not applicable. 

\noindent\emph{Contributions.}
Our first contribution is a clear, mathematically rigorous definition of the fuzzy unreliability metric. The definition is valid for general fuzzy numbers, rather than specific types such as triangular numbers. The definition follows Zadeh's extension principle \cite{zadeh1965fuzzy}, a general approach to apply functions and arithmetic operations on sets to fuzzy numbers.

We then propose a bottom-up algorithm for calculating fuzzy unreliability metric for tree-structured FTs. During the calculation procedure, fuzzy attribution is discretised horizontally and saved as $\alpha$-cut series. Arithmetic operations are performed on these $\alpha$-cut series representing fuzzy numbers. Output of the algorithm is an $\alpha$-cut series approximation of a fuzzy number. This approximate computational technique works for fuzzy numbers that can be expressed as $\alpha$-cut interval and is applicable when performing operations on fuzzy numbers of different types.

Summarized our contributions are:
\begin{enumerate}
\item A rigorous definition of fuzzy unreliability metric;
\item A bottom-up algorithm for computing fuzzy unreliability in tree-like FTs based on $\alpha$-cuts;
\item A counter example showing why a straight forward extension of BDD-based methods for reliability calculations does not work for fuzzy fault trees;
\item An implementation of the algorithm and its illustration on two benchmark fault trees. 

\end{enumerate}

\section{Related work}
\label{sec:related_works}

Fuzzy fault tree analysis has been studied in a wide body of literature and reviewed in many works\cite{mahmood2013FFTA,ruijters2015FTA}. When vagueness arise, uncertain parameters can be described as fuzzy numbers. Each BEs is then equiped with a fuzzy number/possibility. Fuzzy set theory was first introduced in fault tree analysis by Tanaka et al.\cite{tanaka1983fault}. The main idea of using fuzzy theory is to express the ambiguous parameters in triangular, or trapezoidal, or Left-Right type fuzzy numbers, then calculate the fuzzy probability of the top event based on Zadeh's principle as in \cite{tanaka1983fault}, or using $\alpha$-cut fuzzy operators as in\cite{lin1997hybridFTA}, or applying extended algebraic operations on the L-R type fuzzy number as in\cite{singer1990fuzzysetFT}. 

In general, it is challenging to calculate the fuzzy probability of a top event. This is because of the multiplication of fuzzy numbers. Nonlinear programming of the operation is infeasible to implement in practice \cite{DSW1985fuzzy}. Multiplication of trapezoidal or triangular numbers is often done using an estimation \cite{tanaka1983fault} which results in a fuzzy number of the same type as the operands. An approximate analytical algorithm for calculating fuzzy operations is L-R representation\cite{Dubois1979FuzzyRA}. Also, there is an approximate numerical approach \cite{Schmucker1983FuzzySN} that applies the extension principle to the discretised fuzzy numbers resulting in the discrete extended product. This approach, however, lacks generality because the extended operation may not result in convex numbers. Using $\alpha$-cut representation and interval arithmetic, DSW algorithm \cite{DSW1985fuzzy} provides an efficient way to calculate fuzzy operations. We incorporate this method into our BU-algorithm.


\section{Preliminary Theory}\label{sec:fundamentals_fuzzy}

Fuzzy set theory was first introduced by L.A. Zadeh~\cite{zadeh1965fuzzy} in 1965. Since then it has been used to deal with vague problems. We consider fuzzy elements that can have a range of possible values.
For an element $x$ of a set $X$, the extent to which $\mathsf{x}$ can be equal to $x$ is expressed by the \emph{membership degree} of $x$ in $ \mathsf{x}$, which is a value $\mathsf{x}[x] \in [0,1]$. 
By that, $\mathsf{x}[x] = 1$ denotes full membership, while $\mathsf{x}[x] = 0$ denotes no membership.

\begin{figure}
\begin{subfigure}[h]{0.3\linewidth}
\centering
\includegraphics[width=0.7\linewidth]{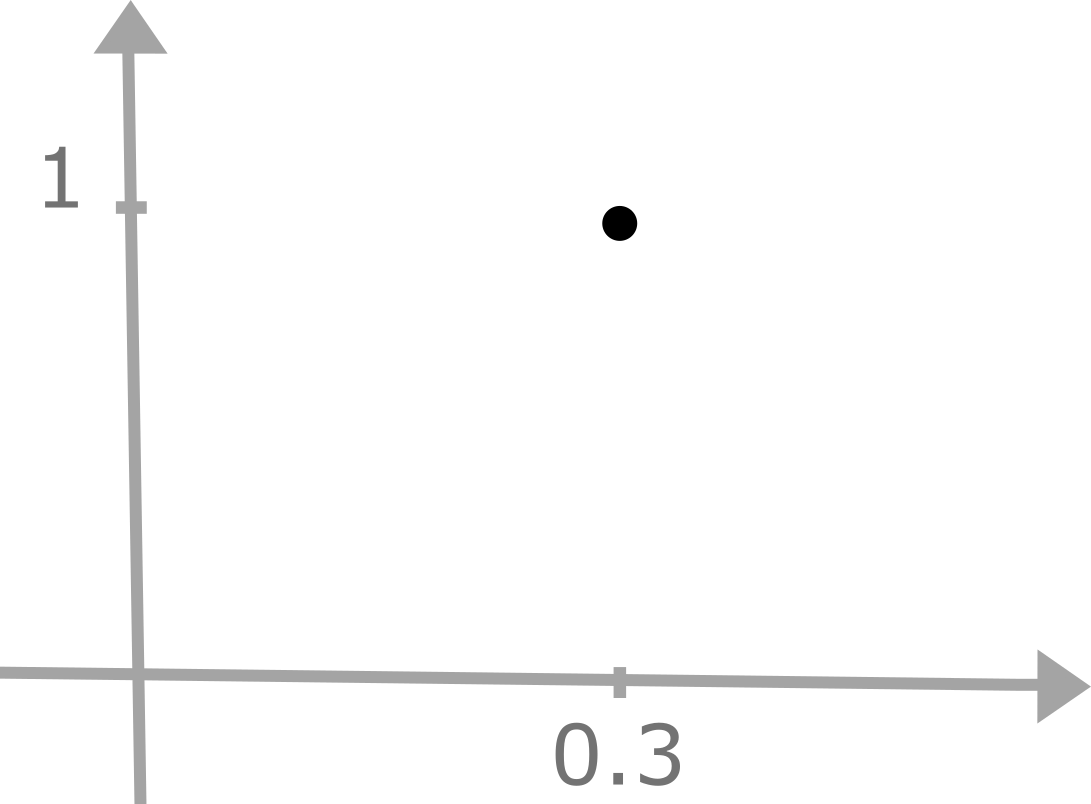}
\caption{}
\label{fig:classic_element}
\end{subfigure}
\hfill
\begin{subfigure}[h]{0.3\linewidth}
\centering
\includegraphics[width=0.7\linewidth]{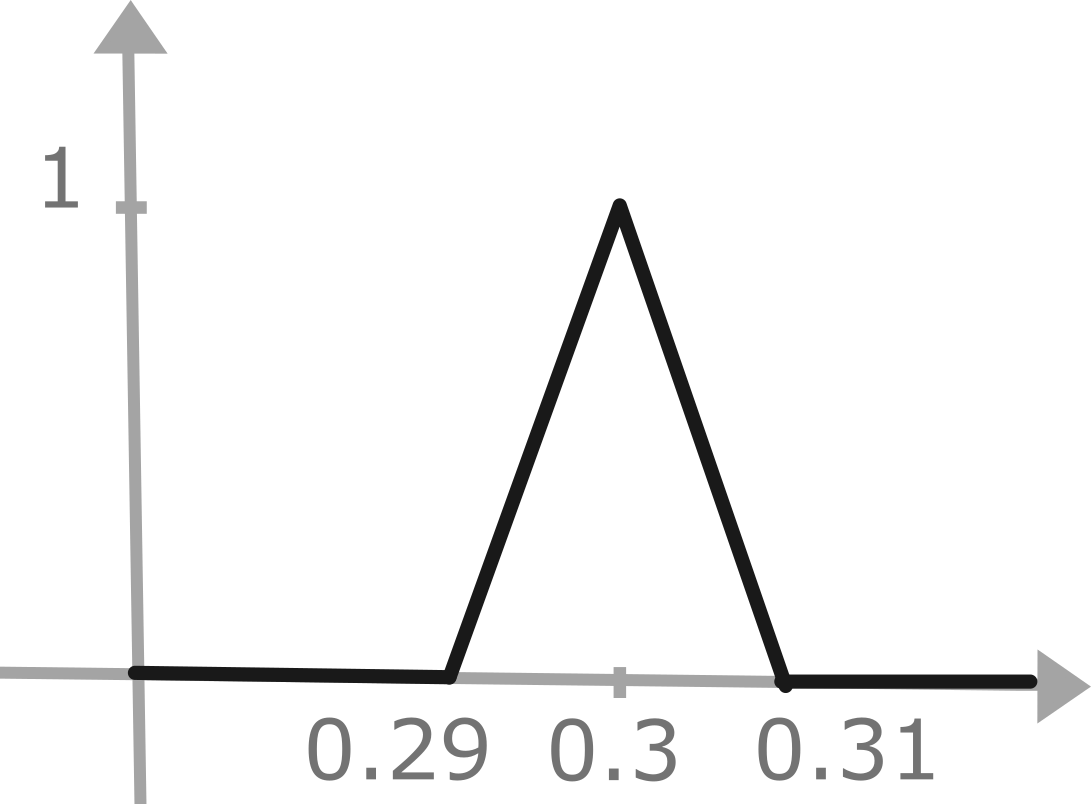}
\caption{}
\label{fig:fuzzy_element}
\end{subfigure}%
\begin{subfigure}[h]{0.4\linewidth}
\centering
\includegraphics[width=0.9\linewidth]{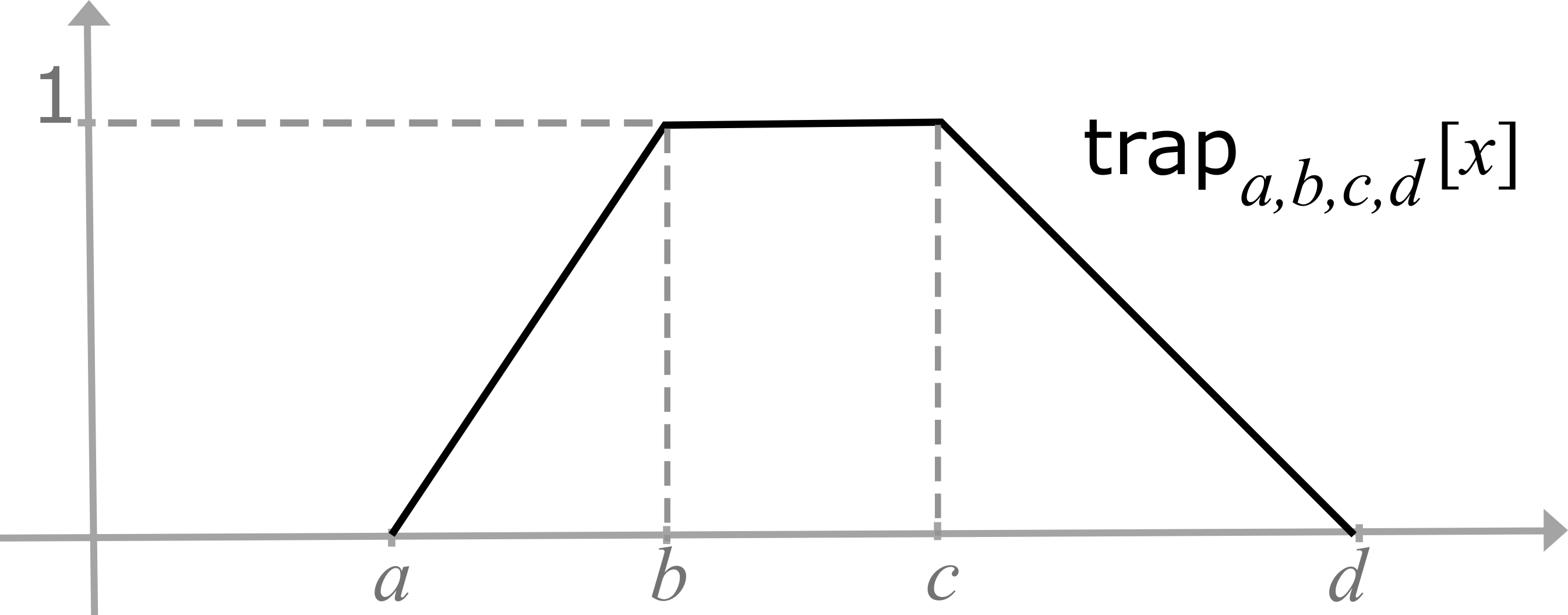}
\caption{}
\label{fig:ex:trapezoidal}
\end{subfigure}%
\caption{ (a) A non-fuzzy, `crisp' element $x=0.3$, (b) a fuzzy element $\mathsf{x}$, expressing uncertainty around 0.3, and (c) a trapezoidal fuzzy number $\mathsf{trap}_{a,b,c,d}$.}
\label{fig:non-fuzzy_vs_fuzzy}
\end{figure}

For instance, the failure probability of a BE may be given as a real number, e.g. $x = 0.3 \in \mathbb{R}$; but often the exact value is not known precisely, and can be somewhere around $0.3$. This can be represented by a fuzzy number $\mathsf{x}\colon \mathbb{R} \rightarrow [0,1]$ which is $0$ everywhere except close to $0.3$, and which has a maximum at $0.3$ (see Fig.~\ref{fig:non-fuzzy_vs_fuzzy}). 






\begin{definition}
Let $X$ be a set. A \emph{fuzzy element} of $X$ is a function $\mathsf{x}\colon X \rightarrow [0,1]$. The set of all fuzzy elements of $X$ is denoted $\mathbf{F}(X) := \{ \mathsf{x} \ | \ \mathsf{x} \colon X \rightarrow [0,1] \}$.
\end{definition}


Our definition of fuzzy element is very general. Many works in the literature restrict the form of the function $\mathsf{x}\colon X \rightarrow [0,1]$ to make computation more convenient, especially for $X = \mathbb{R}$, i.e., for so-called \emph{fuzzy numbers}. Thus there exist triangular, trapezoidal, Gaussian, etc. fuzzy numbers \cite{Jezewski2017theory}.


\begin{example}
For real numbers $a \leq b \leq c \leq d$, the \emph{trapezoidal fuzzy number} $\mathsf{trap}_{a,b,c,d} \in \mathbf{F}(\mathbb{R})$  is defined as (see Fig. \ref{fig:ex:trapezoidal}):

\begin{align*}
    \mathsf{trap}_{a,b,c,d}[x] &= 
    \begin{cases}
        \tfrac{x-a}{b-a}, & \textrm{if } a < x \leq b,\\
       1, & \textrm{if } b<x<c,\\
        \tfrac{d-x}{d-c}, & \textrm{if } c \leq x < d,\\
        0, & \text{otherwise}.
    \end{cases}
\end{align*}
Triangular fuzzy numbers are special cases of trapezoidal numbers, with $b=c$.
Thus, $\mathsf{tri}_{a,b,d} = \mathsf{trap}_{a,b,b,d}$.
Gaussian fuzzy numbers are characterized by their mean $m$ and standard deviation $d$:
\[\mathsf{gauss}_{m,d}[x] = \exp{(\tfrac{-(x-m)^2}{2d^2})}\]
\end{example}
%


For convenience, we  abbreviate $\mathsf{x}$ via a list of membership values $x \mapsto \mathsf{x}[x]$, omitting $x$ for which $\mathsf{x}[x] = 0$. For example, $\mathsf{x} = \{1 \mapsto 0.7,2 \mapsto 0.5\} \in \mathbf{F}(\mathbb{Z})$ 
denotes the fuzzy number with $\mathsf{x}[1] = 0.7$, $\mathsf{x}[2] =0.5$ if $x=2$ and $\mathsf{x}[x] = 0$ everywhere else.

\paragraph{Zadeh's extension principle.}
They key technique to perform arithmetic operations on fuzzy numbers is 
 based on Zadeh's extension principle~\cite{Jezewski2017theory,zadeh1965fuzzy}. 
 To understand this principle, consider a function $f:X \to Y$. Zadeh's extention principle lifts $f$ to a function over fuzzy numbers 
 $\widetilde{f}\colon \mathbf{F}(X)\rightarrow \mathbf{F}(Y)$.
 First assume the special case where $f$ is injective and $f(x)=y$. 
 Then we set $\widetilde{f}(\mathsf{x})[y]= \mathsf{x}[y]$, since the trust we have that for $f(x)$ to be equal to $y$ is the same the trust we have for $x$ to be $y$. 
 For $y\in Y$ without an inverse, i.e. with $f^{-1}(y) = \emptyset$, we set $\widetilde{f}(\mathsf{x})[y]=0$, since  $f$ can never be equal to $y$.
 Now, if $f$ is not injective, then there can be multiple values $x$ with $f(x)=y$. Each of these values $x\in f^{-1}(y)$ have been assigned a fuzzy  value $\mathsf{x}$. Zadeh's extension principle takes the best value from $x\in f^{-1}(y)$, i.e. for which $\mathsf{x}[x]$ is maximal. For functions of multiple arguments this works exactly the same.

\begin{definition}[Zadeh's Extension Principle]\label{def:extension_principle}
Let $f$ be a multiargument function $f:X_1\times X_2\times\dots \times X_n \to Y$. The \emph{Zadeh extension} of $f$ is the function $\widetilde{f}\colon \mathbf{F}(X_1) \times \ldots \times \mathbf{F}(X_n) \rightarrow \mathbf{F}(Y)$ defined as:
\[
\widetilde{f}(\mathsf{x}_1,\ldots,\mathsf{x}_n)[y] = 
\begin{cases}
    \underset{\substack{(x_1,x_2,\dots,x_n) \in \prod_i X_i\colon\\
    f(x_1,x_2,\dots,x_n)=y}}{\sup}\min\limits_{i=1,\dots,n}\ \mathsf{x}_i[x_i], & f^{-1}(y) \neq \emptyset,\\
    0 & f^{-1}(y) = \emptyset.
\end{cases}
\]
\end{definition}


Addition and subtraction operations performed on specific fuzzy numbers can have straightforward formulations. E.g., for two trapezoidal fuzzy numbers we have
\begin{align*}
    \mathsf{trap}_{a_1,a_2,a_3,a_4} \  \widetilde{+} \ \mathsf{trap}_{b_1,b_2,b_3,b_4} &= \mathsf{trap}_{a_1 + b_1, a_2 + b_2, a_3 + b_3, a_4 + b_4},\\
    \mathsf{trap}_{a_1,a_2,a_3,a_4} \ \widetilde{-} \ \mathsf{trap}_{b_1,b_2,b_3,b_4} &= \mathsf{trap}_{a_1 - b_4, a_2 - b_3, a_3 - b_2, a_4 - b_1}.
\end{align*}

The key problem with fuzzy operations is that for many common operations, Zadeh's extension principle destroys the shape of their operant. E.g. 
For example, due to their non-linearity, the quotient of two trapezoidal fuzzy numbers is not trapezoidal; the same is true for their product.

Some works approximate operations by a fuzzy number of the same type as the operations. 
Such approximations yield outcomes that are easy to visualize, but are not exact~\cite{Basiura2015advances}.

\paragraph{$\alpha$-cuts}
Another way to perform arithmetic operations on fuzzy number is 
via their $\alpha$-cuts~\cite{Basiura2015advances}. 
An $\alpha$-cut of $\mathsf{x}^{(\alpha)}$ of 
a fuzzy number $\mathsf{x}\in \mathbf{F}(X)$ yields all values in $x$ with $\mathsf{x}[x] \ge \alpha$, i.e. whose trust is at least $\alpha$.

\begin{definition}\label{def:alpha_cut}
Let $\mathsf{x} \in \mathbf{F}(X)$ and $\alpha
\in [0,1]$. An $\alpha$-cut or $\alpha$-level $\mathsf{x}^{(\alpha)}$ of $\mathsf{x}$ equals 
\begin{equation}\label{eq:alpha_cut}
    \mathsf{x}^{(\alpha)} =\{x \in X \ | \ \mathsf{x}[x] \ge \alpha \}.
\end{equation}
\end{definition}

\begin{figure}[ht]
    \begin{center}
    \includegraphics[width=0.5\textwidth]{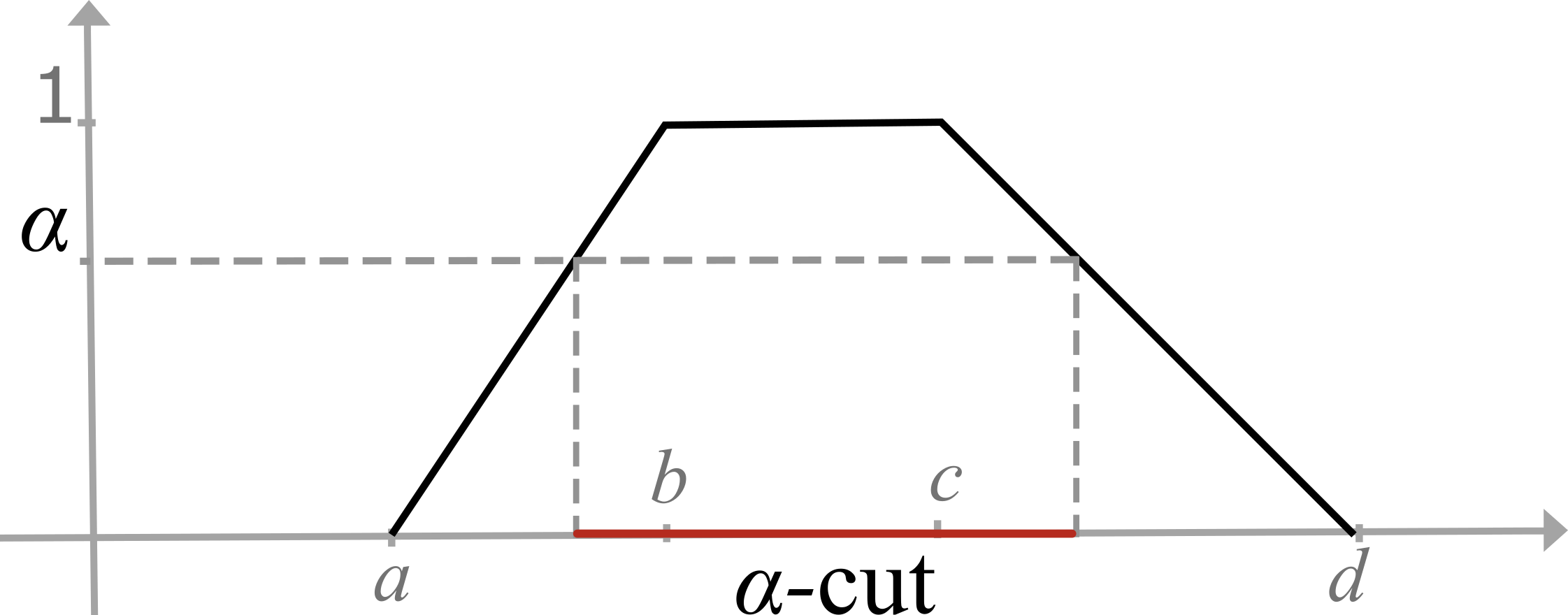}
    \end{center}
    \caption{$\alpha-$cut of trapezoidal fuzzy number $\mathsf{trap}_{a,b,c,d}$.}
    \label{fig:ex:trap_cuts}
\end{figure}

\begin{example}
The $\alpha$-cuts of  trapezoidal and guassian fuzzy number are given by:  

\begin{align*}
\mathsf{trap}^{(\alpha)}_{a,b,c,d} = [(b-a)\alpha + a, \ d - (d-c)\alpha] \\
\mathsf{gauss}_{m,d}^{(\alpha)} = \bigl[m-d\sqrt{-2\ln{\alpha}}, \ m+d\sqrt{-2\ln{\alpha}} \ \bigr] 
\end{align*}

The elements of the set of $\alpha$-cut lie in the red interval shown in Fig. \ref{fig:ex:trap_cuts} 
\end{example}

Suppose we have a fuzzy number $\mathsf{x}$ that is (piecewise) continuous and convex --- i.e., for all $x,y \in \mathbb{R}$ and $\lambda \in \mathbb{R}$ we have $\mathsf{x}[\lambda x+(1-\lambda)y] \ge \min \{\mathsf{x}[x], \mathsf{x}[y]\}$. For such a fuzzy number each $\alpha$-cut $\mathsf{x}^{(\alpha)}$ is a closed interval $[a^{(\alpha)}_1, a^{(\alpha)}_2]$. Furthermore, $\mathsf{x}$ is completely determined by this collection of intervals, as $\alpha$ ranges over $[0,1]$. This means that we have two ways of describing such a fuzzy number: either by the membership function $\mathsf{x}\colon \mathbb{R} \rightarrow [0,1]$, or by the two functions $a_1,a_2\colon [0,1] \rightarrow \mathbb{R}$, that assign to $\alpha$ the endpoints of the $\alpha$-cut of $\mathsf{x}$. The advantage of the second viewpoint is that Zadeh extensions become significantly easier to compute: it turns out that Zadeh extensions of arithmetic operations become interval operations on the $\alpha$-cuts. These operations are defined as follows.


\begin{definition}(Interval operations\cite{ALEFELD2000interval,deBarros2017afirst})\label{def:interval_operations}
    Let $[a_1,a_2]$  and $ [b_1,b_2]$ be two closed intervals on the real line. The arithmetic operations between intervals is defined by
    \begin{enumerate}
        \item $[a_1,a_2]+[b_1,b_2]=[a_1+b_1, a_2+b_2]$
        \item $[a_1,a_2]-[b_1,b_2]=[a_1-b_2, a_2-b_1]$
        \item $[a_1,a_2] \cdot [b_1,b_2]=[\min \{a_1 b_1,a_1 b_2, a_2 b_1, a_2 b_2\}, \max \{a_1 b_1,a_1 b_2, a_2 b_1, a_2 b_2\}]$
    \end{enumerate}
\end{definition}

The following result shows that we can use these interval operations to calculate Zadeh extensions, by describing the resulting fuzzy number in terms of its $\alpha$-cuts. Note that many existing fuzzy number types such as triangular, trapezoidal, Gaussian, satisfy its assumptions.

\begin{lemma}\label{def:binary_operator_alpha_cut}
    Let $\mathsf{x} , \mathsf{y}$ be two piecewise continuous, convex fuzzy numbers, and let $\diamond$ be an arithmetic operation  in $\{+, -, \ \cdot\}$. Then the $\alpha$-cuts of the fuzzy number $\mathsf{x} \widetilde{\diamond} \mathsf{y}$ are given by
    \[ ( \mathsf{x} \ \widetilde{\diamond} \ \mathsf{y} )^{(\alpha)}  = \{x \diamond y | x \in \mathsf{x}^{(\alpha)}, \ y \in \mathsf{x}^{(\alpha)} \}= \mathsf{x}^{(\alpha)} \diamond \mathsf{y}^{(\alpha)}, \quad \forall \alpha \in [0,1]\]
\end{lemma}

\begin{example}
    Consider two triangular fuzzy numbers $\mathsf{tri}_{1,2,3}$, and $\mathsf{tri}_{3,4,6}$. We wish to calculate addition, subtraction, and multiplication of the two fuzzy numbers using their $\alpha$-cuts. We have the equation describing $\alpha$-cuts of $\mathsf{tri}_{1,2,3}$, and $\mathsf{tri}_{3,4,6}$, for any $\alpha \in [0,1]$:
    $\mathsf{tri}^{(\alpha)}_{1,2,3} = [1+\alpha, 3-\alpha]$, and $\mathsf{tri}^{(\alpha)}_{3,4,6} = [3+\alpha, 6-2\alpha]$
    \begin{itemize}
        \item $(\mathsf{tri}_{1,2,3} \ \widetilde{+} \ \mathsf{tri}_{3,4,6})^{(\alpha)} = [4+2\alpha, 9-3\alpha]$
        \item $(\mathsf{tri}_{1,2,3} \ \widetilde{-} \ \mathsf{tri}_{3,4,6})^{(\alpha)} = [3\alpha - 5, -2\alpha]$
    \end{itemize}
    Because the functions are linear, substituting $\alpha=0$ and $\alpha=1$ provides parameters of the resulting triangular membership functions of the sum and subtraction: $\mathsf{tri}_{4,6,9}$, and $\mathsf{tri}_{-5,-2,0}$. Multiplication of fuzzy numbers requires more comments. We need to search for the minimum and maximum among functions: $\alpha^2 +4\alpha+3, -2\alpha^2+4\alpha+6, -\alpha^2+9, 2\alpha^2-12\alpha+18$. Analysing the values of these functions for $\alpha \in [0,1]$, the following interval is obtained
    \begin{itemize}
        \item $(\mathsf{tri}_{1,2,3} \ \widetilde{\cdot} \ \mathsf{tri}_{3,4,6})^{(\alpha)} = [\alpha^2 +4\alpha+3, 2\alpha^2-12\alpha+18]$
    \end{itemize}
    The functions are not linear, thus replacing $\alpha$ with 0 and 1 provides just the values of $x$, for which the membership function takes values 0 and 1 e.g., $(\mathsf{tri}_{1,2,3} \ \widetilde{\cdot} \ \mathsf{tri}_{3,4,6})[3]=0$, $(\mathsf{tri}_{1,2,3} \ \widetilde{\cdot} \ \mathsf{tri}_{3,4,6})[18]=0$ and $(\mathsf{tri}_{1,2,3} \ \widetilde{\cdot} \ \mathsf{tri}_{3,4,6})[8]=1$. To determine the resulting multiplication fuzzy number, we solve two nonlinear functions: $\alpha^2 +4\alpha+3=x$, and $2\alpha^2-12\alpha+18=x$. The solution are $\alpha_{1,2} = -2 \pm \sqrt{1+x}$ for the first equation, and $\alpha_{1,2} = \frac{6 \pm \sqrt{2x}}{2}$ for the second. In the case of the first equation, we choose $-2 + \sqrt{1+x}$ as the membership function for $x\in [3,8]$ because it provides values in the range $[0,1]$. For the second case, the function $\frac{6 - \sqrt{2x}}{2}$ provides values in ranges $[0,1]$ for $x \in [8,18]$. Finally the multiplication is described by:

    \begin{align}
    (\mathsf{tri}_{1,2,3} \ \widetilde{\cdot} \ \mathsf{tri}_{3,4,6})[x] &= 
    \begin{cases}
        -2 + \sqrt{1+x}, & \textrm{if } 3 \leq x \leq 8,\\
        \tfrac{6 - \sqrt{2x}}{2}, & \textrm{if } 8 < x \leq 18,\\
        0, & \text{otherwise}.
    \end{cases}
    \end{align}
\end{example}

\section{Fault trees}\label{sec:FTs}
In this section, we briefly review the terminology of (static) fault trees. A more comprehensive overview can be found, e.g., at \cite{ruijters2015FTA}. 

\begin{definition}\label{def:FTs}~\emph{\cite{lopuhaa2022efficient}}
    A \emph{fault tree} is a tuple $T=(V,E,t)$, where $(V,E)$ is a rooted directed acyclic graph, and $t$ is a map $t\colon V \rightarrow \{\mathtt{BE}, \mathtt{OR}, \mathtt{AND}\}$ such that $t(v) = \mathtt{BE}$ if and only if $v$ is a leaf for all $v \in V$.
\end{definition}

The root of $T$ is denoted $R_T$. For a node $v \in V$, we let $ch(v)$ be the set of all children of $v$. The set of BEs (basic events) is denoted $\text{BE}_T$.

A \emph{safety event} is a set of BEs that occurs simultaneously. We denote such an event by its corresponding binary vector $\vec{f} \in \mathbb{B}^{\text{BE}_T}$. The extent to which a safety event leads to overall system failure is determined by the \emph{structure function}: for a node $v \in V$ and a safety event $\vec{f}$, the structure function $S_T(v,\vec{f})$ is a Boolean stating whether $\vec{f}$ propagates through $v$.

\begin{definition}
    	\label{def:structure_func}
	Let $T$ be a FT. The \emph{structure function} $S_T\colon V \times \mathbb{B}^{\text{BE}_T} \to \mathbb{B}$ of $T$ is defined, for a node $v \in V$ and a safety event $\vec{f}\in \mathbb{B}^{\text{BE}_T}$, by
\begin{align} 
S_T(v,\vec{f}) =&
\begin{cases}
    \bigvee_{w\in ch(v)}S_T(w,\vec{f})  & \parbox{55pt}{if~$t(v)=\mathtt{OR}$,}\\
    \bigwedge_{w\in ch(v)}S_T(w,\vec{f})  & \parbox{55pt}{if~$t(v)=\mathtt{AND}$},\\
    f_v  & \parbox{55pt}{if~$t(v)=\mathtt{BE}$.}
\end{cases}
\end{align}
\end{definition}

A safety event $\vec{f}$ that reaches the root $R_T$ i.e. $S_T(R_T,\vec{f})=1$ is called a \emph{cut set}. The set of all cut sets of $T$ is denoted $\mathcal{C}_T$.



\subsection{Fault tree reliability analysis}

\begin{figure}[t]
\begin{subfigure}{0.5\linewidth}
\centering
\includegraphics[width=0.7\linewidth]{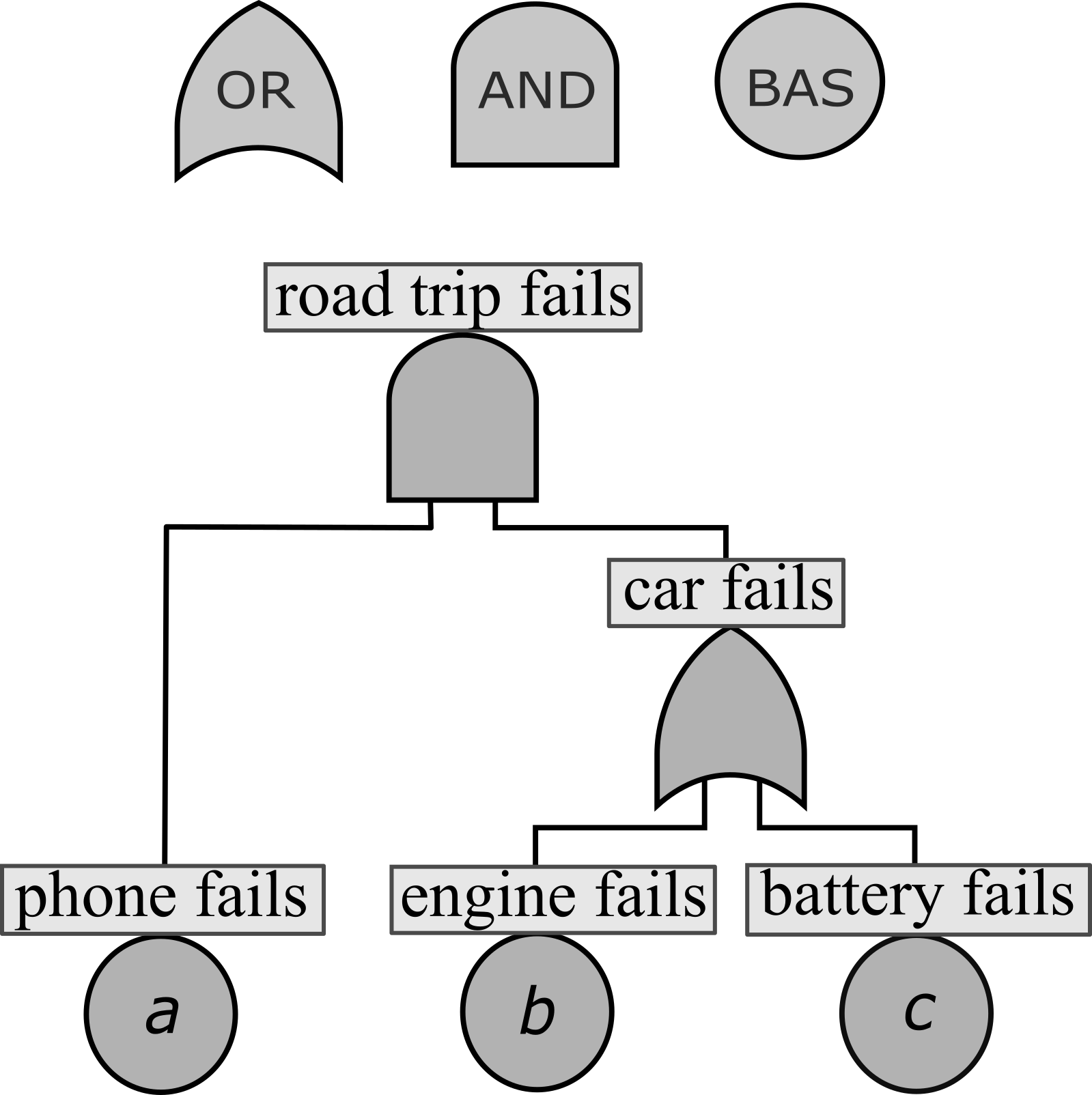}
\caption{}
\label{fig:ex_FT}
\end{subfigure}
\hfill
\begin{subfigure}{0.4\linewidth}
\centering
\includegraphics[width=0.6\linewidth]{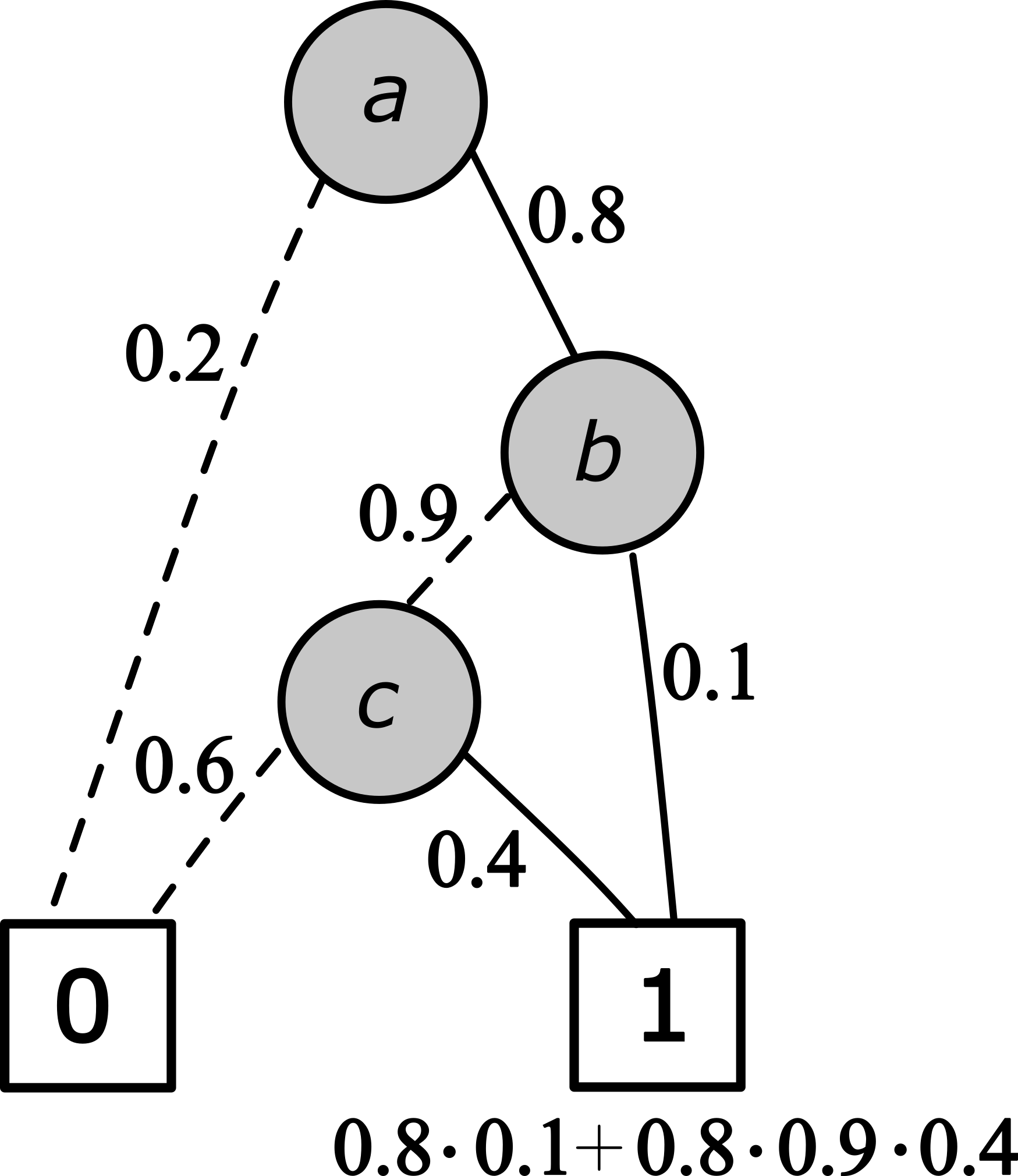}
\caption{}
\label{fig:BDD_ft}
\end{subfigure}%
\caption{A fault tree for a road trip (a), and its BDD representation (b). The road trip fails if both phone and car fail; car fails if either engine or battery fails.}
\label{fig:FT}
\end{figure}

Fault trees are used for quantitative probability analysis as follows: suppose each BE $v$ is assigned a probability $p(v)$, the unreliability of the system is the probability that the system as a whole fails. This is the probability that a random safety event $\vec{F}$ (distributed according to the BE probabilities) propagates to the root. 

\begin{definition}\label{def:unreliability}
    Let $T=(V,E,t,p)$ be a FT. The \emph{unreliability} of $T$ is defined as
    \begin{align}\label{eq:UT}
        U(T) 
             &= \sum_{\vec{f} \in \mathcal{C}_T} \prod_{v=1}^{n} p(v)^{f_v} \cdot \bigl(1-p(v)\bigr)^{(1-f_v)}.
    \end{align}
\end{definition}

\begin{example}\label{ex:compute_unreliability}
    Consider the FT from Fig.\ref{fig:ex_FT}. Let $p(a)=0.8, p(b)=0.1$, and $p(c)=0.4$, then $\vec{f} \in \mathbb{B}^{\text{BE}_T}$ is written as $f_a f_b f_c$. The cut set is represented as $\mathcal{C}_T = \{111,\ 110,\ 101 \}$, and thus the unreliability is
    \begin{align*}
        U(T) &= 0.8 \cdot 0.1 \cdot 0.4 + 0.8 \cdot 0.1 \cdot 0.6 +  0.8 \cdot 0.9 \cdot 0.4 =  0.368.
    \end{align*}
\end{example}

In this definition, the failure probabilities of the BEs are assumed to be independent. This is a standard assumption; any dependence between the BEs is meant to be captured by the FT modeling.

\subsubsection{Bottom-up method} Calculation of \eqref{eq:UT} directly is difficult (in fact, NP-hard \cite{lopuhaazwakenberg2023ftreliability}) because finding the entire set $\mathcal{C}_T$ and calculating each element's probability is copmutationally infeasible for large FTs. Unreliability can instead be computed via a bottom-up method, assigning a probability $p_v$ to each node. For a node $v$ with children $v_1,\ldots,v_n$, we have
\begin{align} 
p_v =&
\begin{cases}
    1 - \prod_{i=1}^{n} (1 - p_{v_i})  & \parbox{55pt}{if~$t(v)=\mathtt{OR}$,}\\
    \prod_{i=1}^{n} p_{v_i}  & \parbox{55pt}{if~$t(v)=\mathtt{AND}$}.
\end{cases} \label{eq:bu}
\end{align}
Then $U(T)$ is calculated as $p_{R_T}$.
\begin{example}\label{ex:compute_probability}
    Consider the FT from Fig.\ref{fig:ex_FT}. Let $p(a)=0.8, p(b)=0.1$, and $p(c)=0.4$. The top failure probability derived using probability laws is    
    \begin{equation*}
        p_{R_T} = p_a \cdot \bigl(1 - (1 - p_b) \cdot (1 - p_c) \bigr) =0.8 \cdot \bigl( 1 - (1 - 0.1) \cdot (1 - 0.4)\bigr) = 0.368.
    \end{equation*}
\end{example}

The bottom-up algorithm is fast, but only works when $T$ is a tree. The reason is that the probability laws in \eqref{eq:bu} only hold when the $p_{v_i}$ are independent probabilities, which will not generally be true if the $v_i$ share children. For DAG-structured FTs, a different approach is used.

\subsubsection{Calculation through BDDs}





Another approach to unreliability calculation involves transforming the structure function of an AT to a binary decision diagram (BDD). BDD are compact representations of Boolean functions; see Fig.~\ref{fig:BDD_ft} for an example. To evaluate a safety event $\vec{f}$, start from the root of the BDD, and at a node labeled $v$, take the solid line if $f_v = 1$, and the dashed line if $f_v = 0$. The label of the reached leaf is the value of $S_T(R_T,\vec{f})$. 

Now $U(T)$ is the probability that one ends up in leaf 1, if we assign probability $p(v)$ to each solid line and $1-p(v)$ to each dashed line from a BDD node labeled $v$. This can be calculated quickly using a bottom-up algorithm on the BDD, and this works even when the original FT is DAG-shaped.
A shortcoming of the method is the worst-case-exponential size of the BDD. The size of BDD is strongly affected by the order of the variable, meanwhile finding the order that minimizes BDD size is NP-hard.   

\section{Fuzzy unreliability}\label{sec:fuzzy_unreliability}
The concept of failure possibility arises in unreliability analysis when the failure probabilities are not exactly known. Instead of having a fixed real value, the failure possibility can be defined by a fuzzy number/ element on interval $[0,1]$, with any shape of membership function. The failure possibility of a basic event $i$ denoted as $\mathsf{p}_i$. We first give an example of how this affects unreliability calculation, before giving the general definition.

\begin{example}\label{ex:fuzzy_unreliability}
    Consider the road trip FT $T$ from Fig.\ref{fig:ex_FT}. Its unreliability function $U_T\colon [0,1]^3 \rightarrow [0,1]$ is given by
    \begin{align*}
    U_T(p_a, p_b, p_c) &= p_ap_bp_c+  p_a(1-p_b)p_c +  p_ap_b(1-p_c)\\
    &= p_ap_c+p_ap_b-p_ap_bp_c.
    \end{align*}
    Now suppose $T$ has failure possibilities $\vec{\mathsf{p}}=(\mathsf{p}_a, \mathsf{p}_b, \mathsf{p}_c)$ given by $\mathsf{p}_a=\{0.5 \mapsto 0.7, 0.8\mapsto 1\}$, $\mathsf{p}_b=\{0.1 \mapsto 1\}$, $\mathsf{p}_c=\{0.4\mapsto 1\}$; that is, $b, c$ have crisp probability values, and $a$ either takes probability $0.5$ or $0.8$ with possibility $0.7$ and $0.1$, respectively. The unreliability of $T$ is then defined to be $\widetilde{U}_T(\mathsf{p}_a, \mathsf{p}_b, \mathsf{p}_c)$, i.e., the Zadeh extension of $U_T$, applied to $\vec{\mathsf{p}}$. Using  Def.\ref{def:extension_principle}, the confidence that this fuzzy unreliability value equal to a number $y\in [0,1]$ is equal to 
\[
\widetilde{U}_T(\vec{\mathsf{p}})[y] = \sup_{\substack{p_a,p_b, p_c \in [0,1]\colon\\ p_ap_c+p_ap_b-p_ap_bp_c  = y}} \min \{ \mathsf{p}_a[p_a], \mathsf{p}_b[p_b], \mathsf{p}_c[p_c] \} 
\]

We notice that $b$ has one single probability value $p_b=0.1$ with $\mathsf{p}_b[p_b] \ne 0$, and $c$ has only $p_c = 0.4$ with $\mathsf{p}_c[p_c] \ne 0$. Substituting those probabilities and their confidences to the expression above, then

\begin{align*}                 
\sup_{\substack{p_a:\\ 0.4p_a+0.1p_a-0.04p_a  = y}}\min \bigl(\mathsf{p}_a[p_a], 1, 1 \bigr)
                 &=\begin{cases}
                        1, & \textrm{ if $y=0.368$},\\
                        0.7, & \textrm{ if $y=0.23$},\\
                        0, & \textrm{ otherwise},
                    \end{cases}
    \end{align*}
so $\widetilde{U}_T(\vec{\mathsf{p}}) = \{0.23 \mapsto 0.7, 0.368 \mapsto 1\}$.
\end{example}


\begin{definition}\label{def:fuzzy_unreliability}
    Let $T$ be a FT. 
    \begin{enumerate}
            \item A \emph{fuzzy attribution} is an element $\vec{\mathsf{p}}$ of $\mathbf{F}([0,1])^{\mathrm{BE}_T}$.

            \item The \emph{fuzzy unreliability value} of $T$ given $\vec{\mathsf{p}}$ is defined as $\widetilde{U}_T(\vec{\mathsf{p}})$, where $\widetilde{U}_T\colon \mathbf{F}([0,1])^{\mathrm{BAS}_T} \rightarrow \mathbf{F}([0,1])$ is the Zadeh extension of the function $U_T$ from Definition \ref{def:unreliability}.
    
    \end{enumerate}
\end{definition}

More concretely, $\widetilde{U}_T(\vec{\mathsf{p}})$ is the fuzzy element of $[0,1]$ defined, for $y\in [0,1]$, by

\begin{align}\label{eq:fuzzy_unreliability}
\widetilde{U}_T(\vec{\mathsf{p}})[y] &= \sup_{\substack{\vec{p} \in [0,1]^{\mathrm{BE}_T}\colon\\ U_T(\vec{p}) = y}} \min_{v \in \mathrm{BE}_T} \mathsf{p}_v[p_v].  \nonumber
\end{align}

\section{Calculation of fuzzy unreliability}

\subsection{Tree-structured FTs}






In general, $\tilde{U}_T(\vec{p})$ is difficult to compute. However, when $T$ is tree-structured, it can be found using a bottom-up algorithm. This algorithm is inspired by the BU algorithm for crisp FT probabilities. When $T$ is a tree, the children of a node $v$ do not have any descendants in common, and as such are independent events. This means that their failure probabilities $\mathsf{p}_v$ can be computed via
\begin{equation} \label{eq:bu}
\mathsf{p}_{v} =
\begin{cases}
    1 - \widetilde{\prod}_{w \in ch(v)} (1 - \mathsf{p}_{w})  & \parbox{55pt}{if~$t(v)=\mathtt{OR}$,}\\
    \widetilde{\prod}_{w \in ch(v)} \mathsf{p}_{w} & \parbox{55pt}{if~$t(v)=\mathtt{AND}$}.
\end{cases} 
\end{equation}

The following result shows the validity of this approach. It is proven analogously to a similar result for attack trees in \cite{dang2024fuzzy}.

\begin{theorem}
Let $T$ be a tree-structured FT, and let $\vec{\mathsf{p}} \in \mathbb{F}([0,1])^{\text{BE}_T}$ be a vector of fuzzy probabilities. Then $\mathsf{p}_{R_T} = \widetilde{U}_T(\vec{\mathsf{p}})$.
\end{theorem}

\begin{example}
    We apply the algorithm to Ex.~\ref{ex:fuzzy_unreliability}. We calculate the $\mathsf{p}_v$ as follows:
    \begin{align*}
    \mathsf{p}_a &= \mathsf{p}_a = \{0.5 \mapsto 0.7, 0.8 \mapsto 1\},\\
    \mathsf{p}_b &= \mathsf{p}_b = \{0.1 \mapsto 1\},\\
    \mathsf{p}_c &= \mathsf{p}_c = \{0.4 \mapsto 1\},\\
    \mathsf{p}_{\texttt{OR}(b,c)} &= 1-(1-\mathsf{p}_b)\tilde{\cdot}(1-\mathsf{p}_c) \\
    &= 1- \{0.9 \mapsto 1\} \tilde{\cdot} \{0.6 \mapsto 1\} = \{0.46 \mapsto 1\},\\
    \mathsf{p}_{R_T} &= \mathsf{p}_a \tilde{\cdot} \mathsf{p}_{\texttt{OR}(b,c)} \\
    &= \{0.5 \mapsto 0.7, 0.8 \mapsto 1\} \tilde{\cdot} \{0.46 \mapsto 1\} = \{0.23 \mapsto 0.7, 0.368 \mapsto 1\}.
    \end{align*}
    Thus we indeed retrieve $\tilde{U}_T(T,\vec{\mathsf{p}}) = \{0.23 \mapsto 0.7, 0.368 \mapsto 1\}$.
\end{example}

\subsection{Efficient computation via $\alpha$-cuts}

While the BU approach calculates $\tilde{U}_T(\vec{\mathsf{p}})$ correctly, it can be difficult to use in practice, due to the fact that the objects $\mathsf{p}_v$ it operates on are functions $[0,1] \rightarrow [0,1]$. Storing such a function in theory would take an infinite amount of space. On the other hand, if one were to approximate this by storing $\mathsf{p}_v[x]$ for only finitely many values of $x$, it is not clear how Zadeh-extended operators such as $\tilde{\cdot}$ are calculated.

To make our algorithm applicable in practice, we choose instead to use $\alpha$-cuts, as we have seen that this allows for efficient computation. First, we choose the (finite) set of $A$ of $\alpha$ we are going to use: we pick the number of cuts $n_{\text{cuts}}$ and take $A = \left\{\tfrac{1}{n_{\text{cuts}}},\tfrac{2}{n_{\text{cuts}}},\cdots,1\right\}$. Then for each BE $v$, we store $\mathsf{p}_v$ as a $n_{\text{cuts}} \times 3$-array whose rows are of the form $(\alpha, a_1(\alpha), a_2(\alpha))$; the interpretation is that $\mathsf{p}_v^{(\alpha)} = [a_1(\alpha),a_2(\alpha)]$. Thus we store each fuzzy number as a set of $n_{\text{cuts}}$ $\alpha$-cuts. The fuzzy multiplication $\tilde{\cdot}$ of \eqref{eq:bu} is now performed level-wise using Lemma \ref{def:binary_operator_alpha_cut}. Thus at each node we need to make $\mathcal{O}(n_{\text{cuts}})$ crisp arithmetic operations, for a total time complexity of $\mathcal{O}(n_{\text{cuts}}\cdot |V|)$.

\subsection{DAG-structured FTs}

When $T$ is not tree-structured but DAG-structured, the BU approach no longer computes $\tilde{U}_T(\vec{\mathsf{p}})$ correctly, because the probabilities of the children of a node may no longer be independent. In fact, this is already true for crisp probability values \cite{ruijters2015FTA}, so it is no surprise that the same holds in the fuzzy setting.

For crisp probabilities the BDD method of Section \ref{sec:FTs} exists
. However, this also does not work in the fuzzy setting. The reason is that the Zadeh-extended arithmetic operators $\tilde{+}$ and $\tilde{\cdot}$ no longer satisfy distributivity, which is crucial in the proof of the BDD method. Thus finding a general algorithm for DAG FTs is considerably more complicated, and we leave this as an open problem.

\begin{figure}[t]
\begin{minipage}[c]{0.4\linewidth}
\centering
\includegraphics[width=0.5\textwidth]{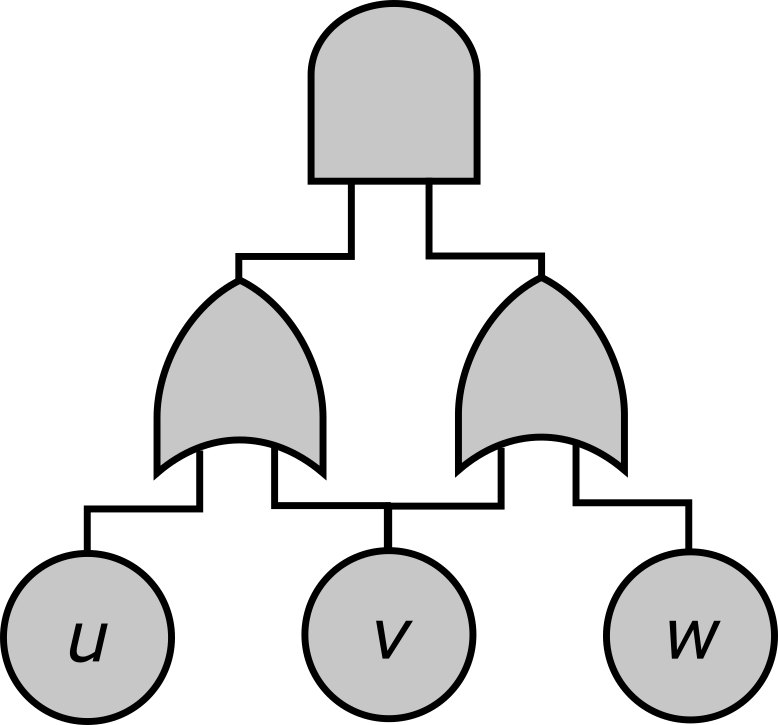}
\caption{A DAG FT}
\label{fig:DAG}
\end{minipage}
\hfill
\begin{minipage}[c]{0.4\linewidth}
\centering
\includegraphics[width=0.6\textwidth]{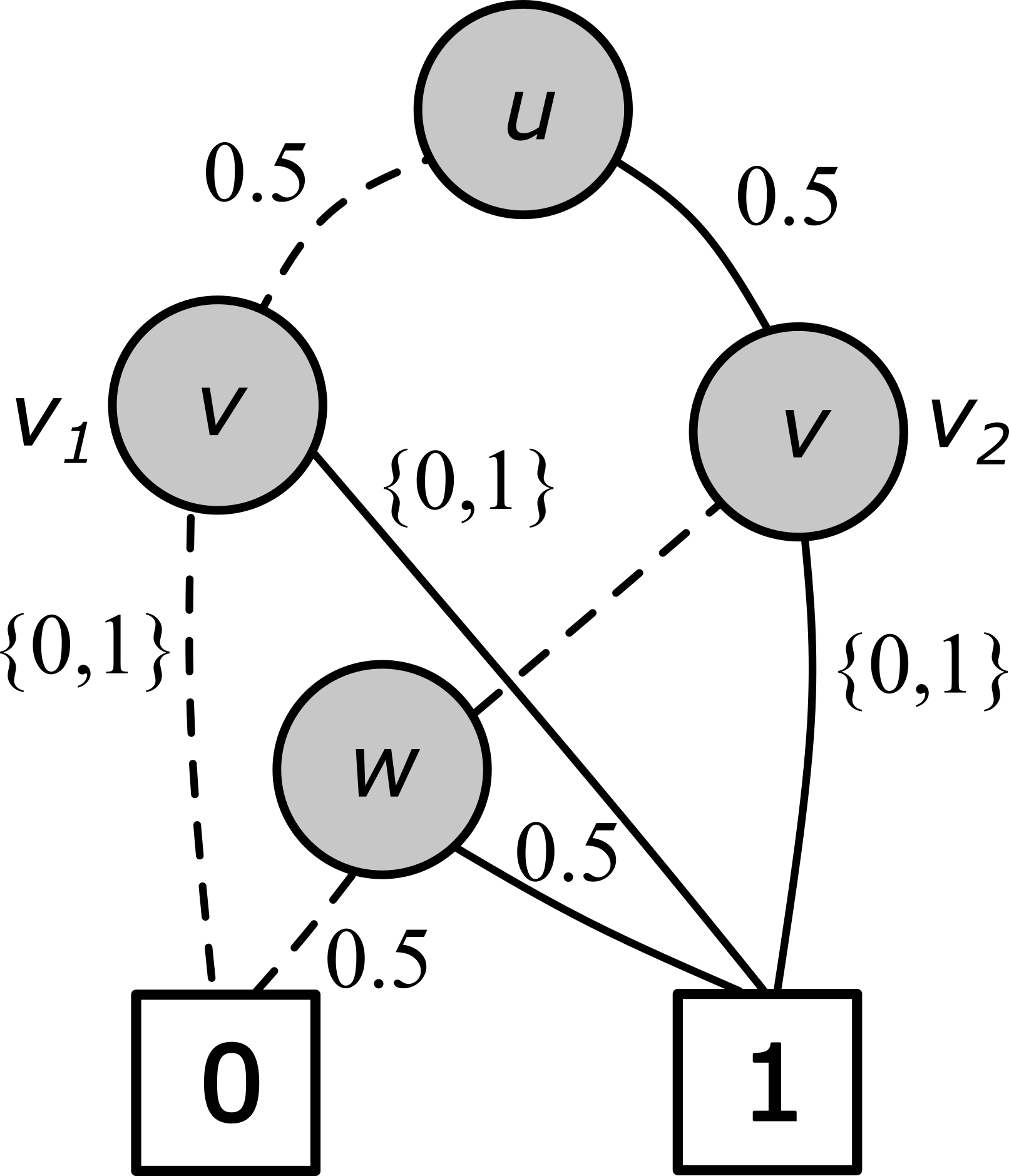}
\caption{BDD calculation}
\label{fig:BDD_cal_FT}
\end{minipage}%
\end{figure}

\begin{example}

    Consider the BDD shown in Fig.\ref{fig:BDD_cal_FT}, with variable order $u<v<w$, and $\mathsf{p}_u = \mathsf{p}_w =  \{0.5 \mapsto 1\}$ and $\mathsf{p}_v = \{0 \mapsto 1, 1 \mapsto 1\}$. We label $v$ as $v_1$ and $v_2$ for distinction in the calculation. Let fuzzy attribution as in the previous example. The fuzzy unreliability of top event is 
    \begin{align*}
        \text{BDD}(v) &= 0.5 \cdot \text{BDD}(v_1) \widetilde{+} 0.5 \cdot \text{BDD}(v_2) \\
        &= 0.5 \cdot \{0 \mapsto 1, 1\mapsto 1\}  \widetilde{+} 0.5 \cdot \{0.5 \mapsto 1, 1\mapsto 1\} \\
        &= \{0 \mapsto 1, 0.5 \mapsto 1\}  \widetilde{+} \{0.25 \mapsto 1, 0.5\mapsto 1\} \\
        &=\{0.25 \mapsto 1, 0.5\mapsto 1, 0.75 \mapsto 1, 1 \mapsto 1\}.
    \end{align*}
    On the other hand, similar to Ex.~\ref{ex:fuzzy_unreliability} one can calculate $\tilde{U}_T(\vec{\mathsf{p}}) = \{0.25 \mapsto 1, 1 \mapsto 1\}$. This shows that the BDD method does not work for fuzzy unreliability.    
\end{example}

\section{Experiments}\label{sec:experiment}
We execute experiments to calculate top event possibility using $\widetilde{BU}$ algorithm. The algorithm is implemented in Python. As we perform computation with fuzzy attribution the system unreliability will be a fuzzy element. We use two benchmark FTs:
\begin{enumerate}
    \item Container Seal Design (CSD)\cite{stamatelatos2002fault} with 6 BEs and 4 ($\mathtt{AND, OR}$) gates. 
    \item Liquid Storage Tank Failure (LSTF)\cite{yazdi2017failure} with 35 BEs and 15 ($\mathtt{AND, OR}$) gates. 
\end{enumerate}


We take the crisp probabilities from \cite{stamatelatos2002fault} and \cite{yazdi2017failure}. In the case of \cite{stamatelatos2002fault}, six BEs $e_1, \dots, e_6$ have crisp probabilities $0.1, 10^{-5}, 10^{-3}, 10^{-3}, 10^{-3}, 10^{-3}$, respectively. In case we want to fuzzify $p_i$ into a triangular possibility, we assign $\mathsf{p}_i=\mathsf{tri}_{0.2\cdot p_i, p_i, 1.8\cdot p_i}$. If we want trapezoidal possibility or Gaussian possibility, we assign $\mathsf{p}_i=\mathsf{trap}_{0.2\cdot p_i, 0.9\cdot p_i, 1.1\cdot p_i, 1.8\cdot p_i}$, or $\mathsf{p}_i=\mathsf{gauss}_{p_i, 0.4\cdot p_i}$, respectively. 

For the experiments, we have fuzzified all crisp BE probability values obtained from the two benchmarks. We use triangular possibility, or trapezoidal possibility, or Gaussian possibility or combination of those three. When combination of the three types arises, fuzzy possibility type for each BEs is assigned as in the table in appendix. BEs possibilities will be horizontally discretised into $n_{cuts} = 100$ $\alpha-$cuts series within the BU algorithm process.

\begin{figure}[t]
\begin{subfigure}[h]{0.5\linewidth}
\includegraphics[width=\linewidth]{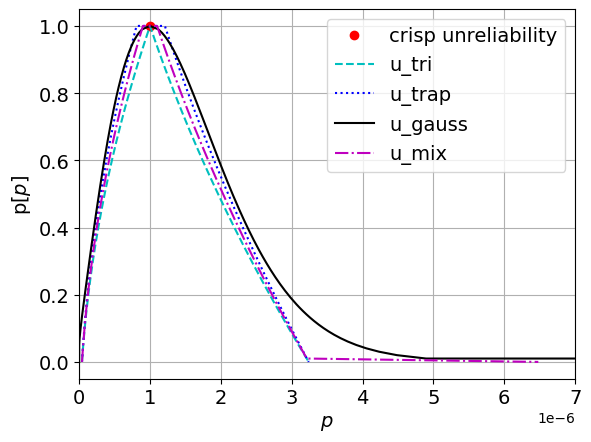}
\caption{CSD}
\end{subfigure}
\hfill
\begin{subfigure}[h]{0.5\linewidth}
\includegraphics[width=\linewidth]{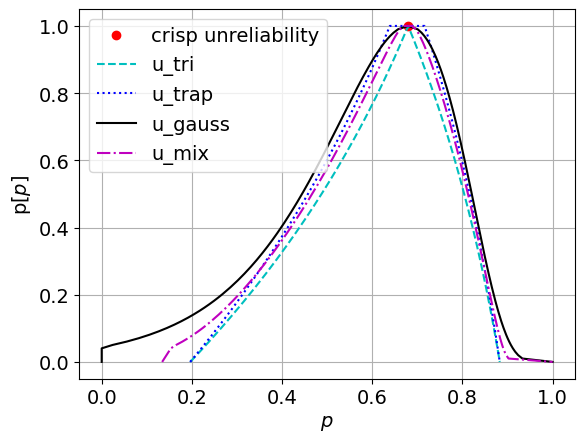}
\caption{LSTF}
\end{subfigure}%
\caption{System fuzzy unreliabilities of  CSD and LSTF FT models}
\label{fig:experiments}
\end{figure}

Figure \ref{fig:experiments} displays the resulting fuzzy unreliabilities of the systems CSD and LSTF, respectively. In each subfigure, we plot the crisp unreliability obtained from crisp BEs values (crisp unreliability), fuzzy unreliability obtained from fuzzy BEs triangular possibility ($\mathsf{u\_tri}$), trapezoidal possibility ($\mathsf{u\_trap}$), Gaussian possibility ($\mathsf{u\_gauss}$), and the their combination ($\mathsf{u\_mix}$). We observe that the system unreliability is not the same type as the operands BEs possibilities. This is obvious since calculation of unreliability requires fuzzy multiplications which is a nonlinear operation. The method overcomes the complexity of performing exact fuzzy multiplication \cite{tanaka1983fault} or using nonlinear programming \cite{DSW1985fuzzy}. In addition, it provides more precise solution than the product estimations of fuzzy numbers and the conventional discretisation approach \cite{Schmucker1983FuzzySN}. 
\section{Conclusion and future work}
In this paper, we define a mathematical formulation for deriving fuzzy unreliability values for FTs. The definition is explicit and generic for general fuzzy attribution. 
We also introduce an efficient algorithm to calculate FT fuzzy unreliability metric. The algorithm works for tree-like structure models with any type of fuzzy attribute that can be expressed as $\alpha$-cut intervals. Experiments on practical models show that the algorithm preserves the nonlinear property of the resulting fuzzy number.  

In the future, we want to develop an algorithm for fuzzy unreliability computation on DAG FTs. Another avenue for future research is to create an algorithm for probabilistic FT analysis in continuous time where each BE is assigned a continuous probability distribution.

%
%
%
\bibliographystyle{splncs04}
\bibliography{references}

\end{document}